\documentclass[preprint,10pt]{elsarticle}

\usepackage{amssymb}
\usepackage{amsmath}
\usepackage{hyperref}
\usepackage{xcolor}
\usepackage{changepage}
\usepackage{enumitem}
\usepackage{multirow}

\journal{}

\begin{document}

\begin{frontmatter}


\author[label1]{Fuzhang Li\corref{cor1}}
\ead{stoneLi20@163.com}
\cortext[cor1]{Corresponding author}
\affiliation[label1]{organization={School of Electronic Engineering, Guangxi University of Science and Technology},
	city={Liuzhou},
	postcode={545006}, 
	state={Guangxi},
	country={China}}

\title{UHNet: An Ultra-Lightweight and High-Speed Edge Detection Network}

%

\author[label2]{Chuan Lin} 
\affiliation[label2]{organization={School of Automation, Guangxi University of Science and Technology},
	city={Liuzhou},
	postcode={545006}, 
	state={Guangxi},
	country={China}}

\begin{abstract}
Edge detection is crucial in medical image processing, enabling precise extraction of structural information to support lesion identification and image analysis. Traditional edge detection models typically rely on complex Convolutional Neural Networks and Vision Transformer architectures. Due to their numerous parameters and high computational demands, these models are limited in their application on resource-constrained devices. This paper presents an ultra-lightweight edge detection model (UHNet), characterized by its minimal parameter count, rapid computation speed, negligible of pre-training costs, and commendable performance. UHNet boasts impressive performance metrics with 42.3k parameters, 166 FPS, and 0.79G FLOPs. By employing an innovative feature extraction module and optimized residual connection method, UHNet significantly reduces model complexity and computational requirements. Additionally, a lightweight feature fusion strategy is explored, enhancing detection accuracy. Experimental results on the BSDS500, NYUD, and BIPED datasets validate that UHNet achieves remarkable edge detection performance while maintaining high efficiency. This work not only provides new insights into the design of lightweight edge detection models but also demonstrates the potential and application prospects of the UHNet model in engineering applications such as medical image processing. The codes are available at \href{https://github.com/stoneLi20cv/UHNet}{https://github.com/stoneLi20cv/UHNet}.
\end{abstract}




\end{frontmatter}



\section{Introduction}
\label{sec1}
Edge detection, a crucial foundational technique in computer vision, has had a profound impact on various medical image processing domains, such as X-rays, CT scans, and MRI images. These images contain rich structural information, and edges are significant manifestations of these structures. Accurate edge detection not only aids in the clear identification of pathological regions but also provides robust support for subsequent image analysis and diagnosis. Therefore, lightweight, fast, and high-performance edge detection models present increasingly complex challenges in practical deployment scenarios.

Specifically, convolutional neural networks (CNNs) \cite{gu2018recent}, as a mainstream deep learning architecture, have seen numerous derivative algorithms applied to edge detection tasks. Traditional algorithms \cite{xie2015holistically,kokkinos2015pushing,maninis2016convolutional,yang2016object,liao2017deep,liu2017richer,wang2017deep,xu2017learning,yu2017casenet,deng2018learning,he2019bi,kelm2019object,wang2018deep,yang2019contourgan,cao2020learning,lin2020lateral,linsley2020recurrent,li2021bi,gao2021bottom,wibisono2021fined,pu2021rindnet,huan2021unmixing,bao2022bidirectional,lin2022bio,zhou2023learning,elharrouss2023refined,ye2024diffusionedge,xian2024feature} based on pre-trained models use series backbone networks like VGG \cite{simonyan2014very} and ResNet \cite{he2016deep} (often referred to as encoder networks), focusing on and developing decoder network structures. This results in a series of large-parameter, non-lightweight, and computationally intensive edge detection models, with most models operating at relatively slow speeds. Consequently, this leads to substantial GPU resource consumption and training time, making it difficult to deploy these models on low-computation edge devices.

Furthermore, Vision Transformer (ViT) \cite{han2022survey,khan2022transformers}, leveraging self-attention mechanisms to model long-range dependencies in images, excel in understanding global context. Despite their superior performance over CNNs in some tasks, their higher computational complexity and memory requirements often limit their use in real-time applications or resource-constrained environments.

Thus, achieving both efficient training and inference while maintaining considerable detection performance is a challenging problem in edge detection tasks. Exploring this issue is essential for cost-effectively deploying efficient network models in real-world applications. A direct solution is designing lightweight detection models based on CNNs \cite{gu2018recent}. Lightweight network design, as an effective solution, provides new insights for reducing model complexity and computational load by optimizing network structures and reducing parameter numbers. However, despite some progress in lightweight network design for edge detection tasks, there remains a trade-off between achieving ultra-high speed and maintaining high accuracy. Given the complexity and diversity of images, how to achieve high-precision edge detection while ensuring ultra-high processing speed is the core issue this paper focuses on.

This paper proposes an ultra-lightweight network model with minimal parameters, extremely fast computation speed, no pre-training cost, and considerable performance, aimed at detecting target edges. Our innovative work primarily focuses on the following four aspects:

\begin{enumerate}[label=(\arabic*),labelindent=1.6em,leftmargin=*]  
\item Proposing an ultra-lightweight feature extraction module, PDDP block. This module evolves from the Bottleneck structure in ResNet \cite{he2016deep}, extracting and integrating target edge features in images with few parameters and high speed.

\item Replacing the original 1x1 convolution for channel transformation with max pooling (MaxPool) and average pooling (AvgPool) operations in different stages of the backbone network, further reducing computation while enhancing feature diversity.

\item Exploring lightweight fusion methods for features between different stages, improving edge detection accuracy through effective feature fusion strategies with fewer parameters compared to other fusion methods.

\item Experiments demonstrate that the proposed ultra-lightweight network model (UHNet) with minimal parameters (42.3k), high speed (166 FPS), and low FLOPs (0.79G) shows strong competitiveness across multiple public datasets.
\end{enumerate} 

The rest of this paper is structured as follows: Section \ref{sec2} explains the work related to edge detection. In Section \ref{sec3}, we describe our proposed method in detail. In Section \ref{sec4}, we conduct detailed experimental verification and comparative analysis of the proposed method on three datasets: BSDS500, NYUD, and BIPED. In Section \ref{sec5}, we summarize the entire paper and discuss directions worthy of further exploration in this paper.

\section{Related Work}
\label{sec2}
With the development of deep learning in computer vision, deep learning-based edge detection methods have achieved significant progress. Xie et al. proposed the first CNN-based edge detection model, HED \cite{xie2015holistically}, using VGG16 \cite{simonyan2014very} as the backbone network, demonstrating strong capabilities in extracting target edges from RGB images. Subsequently, numerous edge detection methods emerged, employing VGG \cite{simonyan2014very}, or ResNet \cite{he2016deep} as backbone networks. These methods \cite{xie2015holistically,kokkinos2015pushing,maninis2016convolutional,yang2016object,liao2017deep,liu2017richer,wang2017deep,xu2017learning,yu2017casenet,deng2018learning,he2019bi,kelm2019object,wang2018deep,yang2019contourgan,cao2020learning,lin2020lateral,linsley2020recurrent,li2021bi,gao2021bottom,wibisono2021fined,pu2021rindnet,huan2021unmixing,bao2022bidirectional,lin2022bio,zhou2023learning,elharrouss2023refined,ye2024diffusionedge,xian2024feature} use transfer learning, pre-training on the ImageNet dataset, followed by fine-tuning on specialized edge detection datasets to further enhance performance. Compared to CNN-based detection models, edge detection methods based on Transformer architectures, like DPED \cite{chen2022dped} and EDTER \cite{pu2022edter}, though performing well, have large parameter numbers leading to higher computational resource demands, posing deployment challenges and cost concerns in practical applications.

Addressing the challenges posed by large models, designing lightweight architectures becomes key to improving edge detection efficiency. Some existing studies \cite{tang2019learning,su2021pixel,soria2023dense,luo2023blednet,pang2024bio} have optimized encoding and decoding network structures or introduced learnable differential convolution operators to reduce model complexity and computational requirements while maintaining detection accuracy. PiDiNet \cite{su2021pixel} proposed an innovative differential convolution operator, dynamically adjusting convolution kernel parameters to better capture edge features, achieving high-performance edge detection under lightweight conditions with a novel network architecture. DexiNed \cite{soria2023dense} abandoned the traditional paradigm relying on large-scale dataset pre-training, demonstrating excellent detection performance directly from edge detection datasets through carefully designed network structures and training strategies, offering new insights into lightweight edge detection model design. Inspired by biological vision pathways, research like LNRFM \cite{tang2019learning}, BLEDNet \cite{luo2023blednet} and XYW-Net \cite{pang2024bio} constructed lightweight and efficient edge detection network models by simulating hierarchical structures of the visual system, reducing computational resource consumption and enhancing adaptability to complex scenes.

However, existing lightweight models commonly face two key issues: 1) The network model design heavily relies on valuable personal experience; 2) Model structures are fixed and single, lacking flexibility, making it difficult to directly transfer and extend to other research tasks. Addressing these issues, as shown in PiDiNet's related experiments \cite{su2021pixel}, differential convolution operators do not always perform positively in all convolution layers. Similarly, models such as LNRFM \cite{tang2019learning}, BLEDNet \cite{luo2023blednet} and XYW-Net \cite{pang2024bio}, which are constructed by simulating biological visual mechanisms, also rely on a deep understanding of the information processing processes in the visual system. However, the performance of network models simply constructed by simulating the physiological mechanisms of the visual system may even be inferior to traditional methods such as HED \cite{xie2015holistically}. Moreover, fixed and single network structures, like LNRFM \cite{tang2019learning}, DexiNed \cite{soria2023dense}, BLEDNet \cite{luo2023blednet}, and XYW-Net \cite{pang2024bio}, cannot form diverse, general-purpose foundational models like VGG \cite{simonyan2014very} and ResNet \cite{he2016deep} series. Unlike these models, we prioritize simplicity, transferability, and scalability in network design, and proposing an ultra-lightweight network model (UHNet) with minimal parameters, high processing speed, and considerable performance, aiming for efficient detection and acquisition of target edges in images.

\section{Method}
\label{sec3}

\subsection{PDDP Block}
\label{subsec31}

\begin{figure}[!ht]
	\centering
	\includegraphics[width=0.6\textwidth]{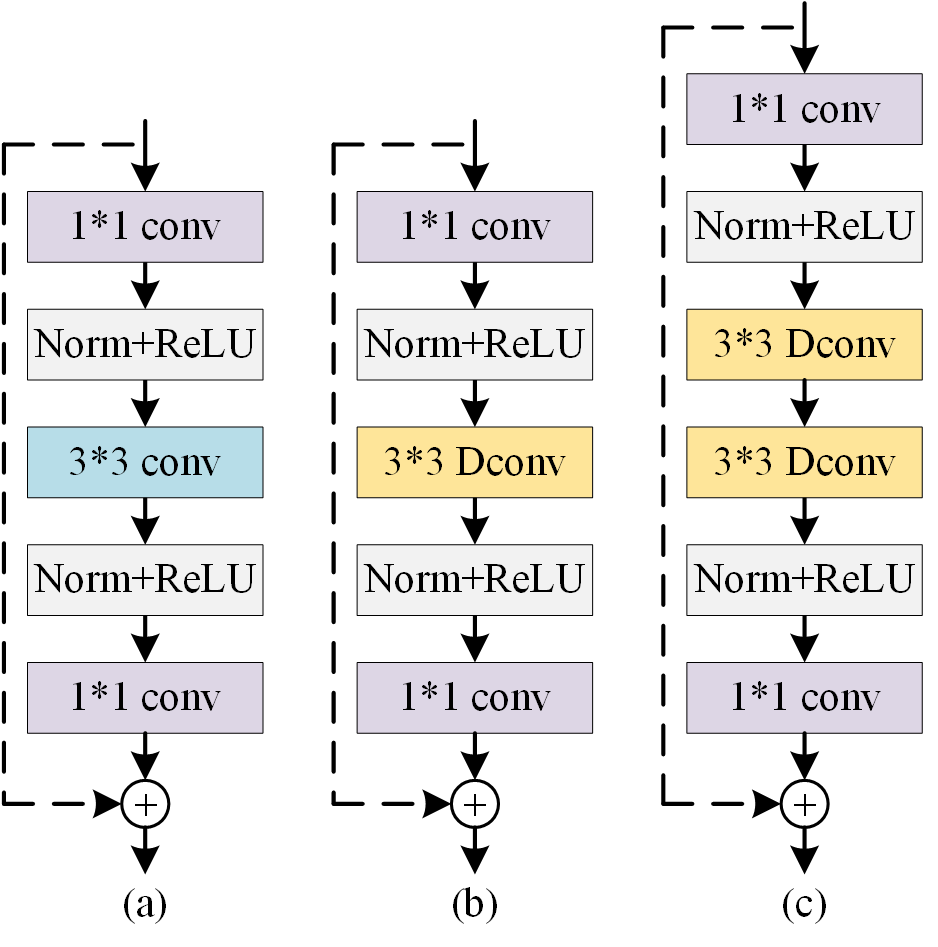}
	\caption{(a) Bottleneck structure in ResNet \cite{he2016deep}; (b) Lightweight Bottleneck structure; (c) PDDP block.}\label{fig1}
\end{figure}

A common lightweight strategy is using Depthwise Separable Convolution to replace standard convolution operations. This convolution method decomposes standard convolution into depthwise and pointwise convolutions, significantly reducing computational load and parameter count. Based on this strategy, the 3x3 standard convolution kernel in the Bottleneck structure of ResNet series networks \cite{he2016deep} (Fig. \ref{fig1}.(a)) is replaced by the depthwise convolution in Depthwise Separable Convolution, resulting in a lightweight Bottleneck structure (Fig. \ref{fig1}.(b)).

The receptive field is particularly important in convolutional neural networks as it directly relates to feature extraction effectiveness. With the increase in network layers, deeper neurons can see larger input areas, capturing more contextual information and improving the model's representation and generalization capabilities. In the lightweight Bottleneck structure (Fig. \ref{fig1}.(b)), the receptive field size is 3x3. To capture more contextual information, we consider increasing the receptive field size by adding another depthwise convolution layer, slightly increasing parameters (Fig. \ref{fig1}.(c)). This structure is called a PDDP block.

Assuming the number of input channels is 32 and the number of output channels in the final 1x1 convolution layer is 64 (other convolution layers have 32 output channels), and ignoring Norm+ReLU layers, the parameter counts for the three structures in Fig. \ref{fig1} are: for the Bottleneck structure in ResNet \cite{he2016deep}, the parameter count is 12288; for the lightweight structure (Fig. \ref{fig1}(b)), the parameter count is 3360; and for the structure with an added depthwise convolution layer to enlarge the receptive field (Fig. \ref{fig1}(c)), the parameter count is 3648. Increasing the receptive field by adding a small number of parameters can improve detection performance while minimally impacting the network's processing speed. We believe this is significant for detecting and acquiring edge information of targets in images.

\subsection{UHNet Architecture}
\label{subsec32}

\begin{figure}[!ht]
	\centering
	\includegraphics[width=0.6\textwidth]{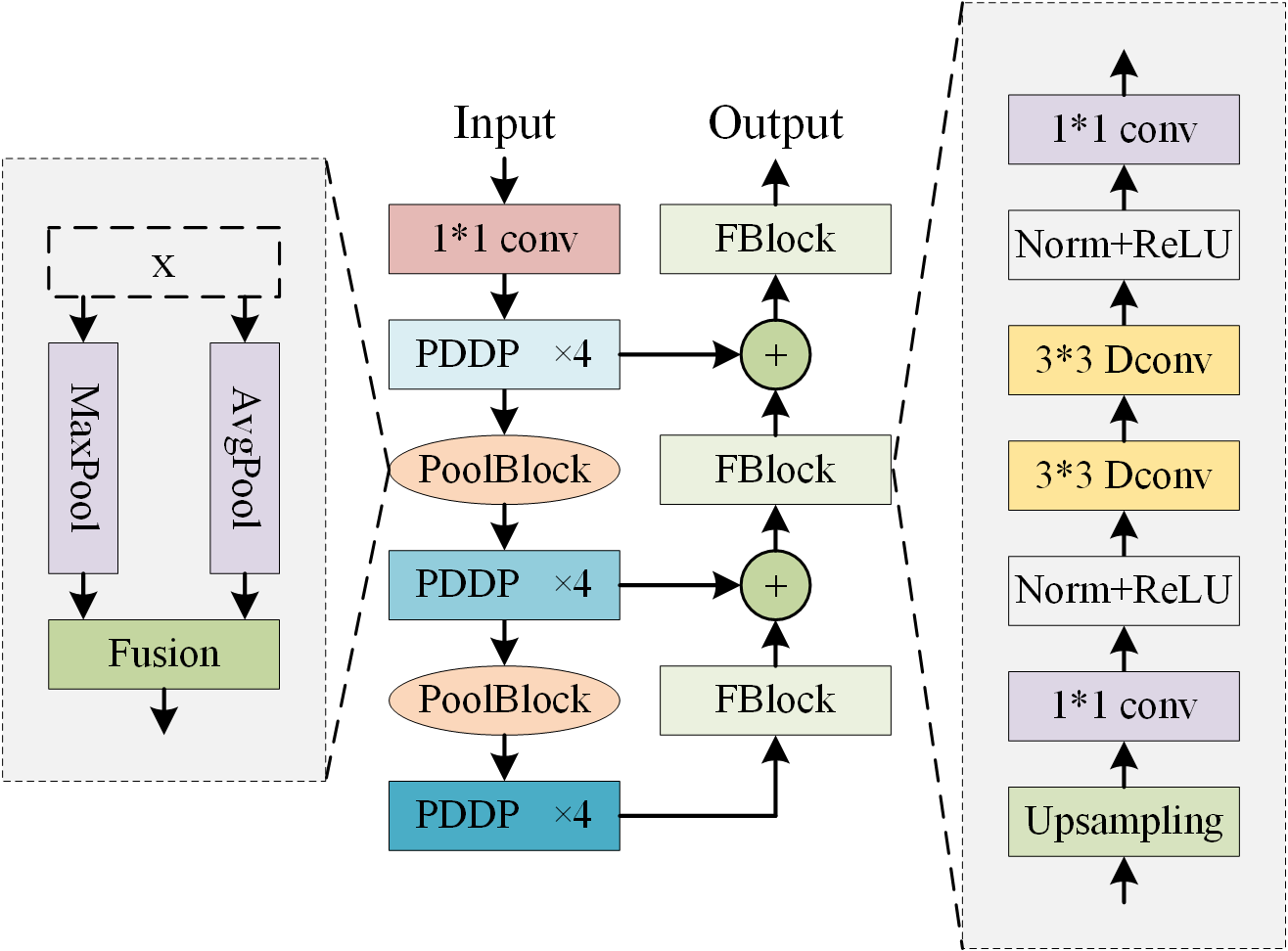}
	\caption{UHNet Architecture.}\label{fig2}
\end{figure}

Fig. \ref{fig2} shows the proposed model structure, mainly composed of the backbone network on the left and the feature decoding network on the right. The backbone network is divided into three stages, separated by PoolBlock between every two stages. Each stage contains four consecutive PDDP blocks, with the first 1x1 convolution layer in the first stage responsible for initial channel transformation of the input image.

In the PoolBlock, the Fusion layer decides the feature fusion method based on the channel numbers of two adjacent stages. Specifically, there are two cases: 1) when the channel numbers of the two adjacent stages are the same, the Fusion layer fuses features by addition; 2) when the channel number of the latter stage is twice that of the former stage, the Fusion layer fuses features by concatenation.

In the feature decoding network, the deep features of the latter stage are first processed by the FBlock, then directly added to the features of the former stage. This process is sequentially applied to three different stages, fusing the output features of each stage to finally obtain the edge detection output.

\section{Experiment}
\label{sec4}

\subsection{Experimental Details}
\label{subsec41}
We evaluated our model using three widely adopted public datasets: BSDS500 \cite{arbelaez2010contour}, NYUD \cite{silberman2012indoor}, and BIPED \cite{poma2020dense}.

The BSDS500 dataset \cite{arbelaez2010contour} comprises 500 images, divided into 200 training images, 100 validation images, and 200 testing images. Each image is annotated by 4 to 9 annotators to ensure accuracy and diversity. To enhance the model's generalization ability, we followed the data augmentation methods from \cite{xie2015holistically,su2021pixel,pang2024bio}, applying flipping, scaling, and rotation to the training images, expanding the training set by 96 times. Additionally, to further enrich the training data's diversity and quantity, we integrated the PASCAL VOC dataset \cite{mottaghi2014role}, which includes 10K labeled images, and augmented it through flipping to 20K images. Ultimately, we obtained a new dataset, BSDS-VOC, containing a total of 48,548 images, providing a richer and more comprehensive data foundation for training edge detection models.

The NYUD dataset \cite{silberman2012indoor} includes 1449 pairs of aligned RGB and depth images, all of which have been densely annotated. The dataset is split into 381 training images, 414 validation images, and 654 testing images. For data augmentation, we merged the training and validation sets and applied flipping (2x), scaling (3x), and rotation (4x) to obtain a more diverse training subset.

The BIPED dataset \cite{poma2020dense} consists of 250 outdoor images with a resolution of 1280×720, each with expert-provided edge annotations. Following the method \cite{poma2020dense,su2021pixel,pang2024bio}, we used 200 images for training and the remaining 50 for testing.

To evaluate model performance, we employed metrics such as Optimal Dataset Scale (ODS), Optimal Image Scale (OIS), and Average Precision (AP) to comprehensively assess the model's accuracy and effectiveness. Additionally, to analyze the model's computational efficiency and size, we introduced the concepts of Floating Point Operations (FLOP) and parameter count as evaluation metrics. The number of parameters directly reflects the model size, while FLOP measures the computational workload during data processing. We also considered the model's Frames Per Second (FPS) as an essential performance metric to evaluate overall efficiency.

We validated the proposed algorithm using the PaddlePaddle deep learning framework \cite{ma2019paddlepaddle} on a computer with 32GB RAM, an NVIDIA GeForce RTX 4090 D 24GB GPU, and an Intel 12th Gen Core i5-12600KF CPU. The parameter settings included the AdamW optimizer \cite{loshchilov2017decoupled}, 15 iterations, a learning rate (lr) of 0.001, a batch size of 1, and a cross-entropy loss function. For the BSDS500 \cite{arbelaez2010contour} and BIPED \cite{poma2020dense} datasets, the maximum allowable error between predicted results and ground truth during Non-Maximum Suppression (NMS) was set to 0.0075. For the NYUD \cite{silberman2012indoor} dataset, this value was set to 0.011.

\subsection{Ablation Study}
\label{subsec42}
We conducted a comprehensive ablation study and analysis of the proposed UHNet on the BSDS-VOC dataset. Notably, in all ablation studies, the number of channels in all three stages was set to 32.

\begin{table}[!ht]
	\centering
	\caption{Performance analysis of different structures. Ns describes the number of standard convolution kernels, Nd describes the number of depthwise convolution kernels.}
	\begin{tabular}{ccccccc}
		\hline
		\textbf{Architecture} & \textbf{Ns} & \textbf{Nd} & \textbf{Params} & \textbf{ODS} & \textbf{OIS} & \textbf{AP} \\ \hline
		RB1 & 1 & 0 & 11.3k & .770 & .791 & .823  \\ 
		RB2 & 2 & 0 & 20.5k & .780 & .802 & .834  \\ 
		LB & 0 & 1 & 2.3k & .779 & .801 & .835  \\ 
		PDDP & 0 & 2 & 2.6k & .784 & .804 & .840  \\ \hline
	\end{tabular}
	\label{table1}
\end{table}

\begin{figure}[!ht]
\centering
\includegraphics[width=0.6\textwidth]{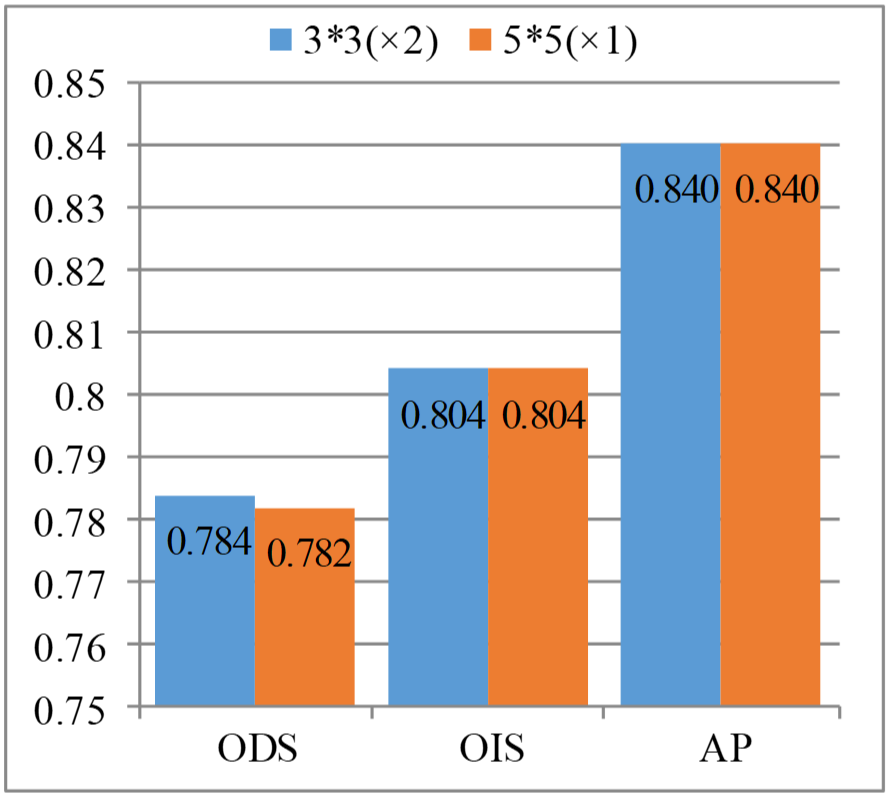}
\caption{Performance comparison between two 3×3 depthwise convolution kernels and one 5×5 depthwise convolution kernel in the PDDP block.}\label{fig3}
\end{figure}

As shown in Table \ref{table1} and Fig. \ref{fig3}, we verified the impact of the ResNet Bottleneck structure \cite{he2016deep} (RB), Lightweight Bottleneck (LB), and PDDP block (PDDP) on model performance through relevant experiments. In this experiment, we focused on three performance aspects: 1) convolution type (standard convolution vs. depthwise convolution); 2) the number of convolution kernels; 3) convolution kernel size. We primarily considered four performance metrics: parameter count (Params), ODS, OIS, and AP, to conduct experimental verification, comparison, and analysis. The results showed that, without additional pre-training, the performance of a single depthwise convolution (LB) significantly outperformed a single standard convolution (RB1) and was almost equivalent to two standard convolutions (RB2). However, the parameter count of LB was only 20.4\% of RB1 and 11.2\% of RB2. Adding another depthwise convolution layer to the LB to form the PDDP structure resulted in improved ODS, OIS, and AP performance, as shown in Table \ref{table1}, with only a 0.3k increase in parameters. We further increased the depthwise convolution kernel size from 3×3 to 5×5 on the LB to verify its impact on model performance. As shown in Fig. \ref{fig3}, when the depthwise convolution kernel was 5×5 (single), the ODS was 0.782, lower than the PDDP's 0.784, with a 0.2k increase in parameter count to 2.8k. This finding demonstrates the effectiveness and potential of the PDDP block in lightweight network design.

\begin{figure}[!ht]
	\centering
	\includegraphics[width=0.6\textwidth]{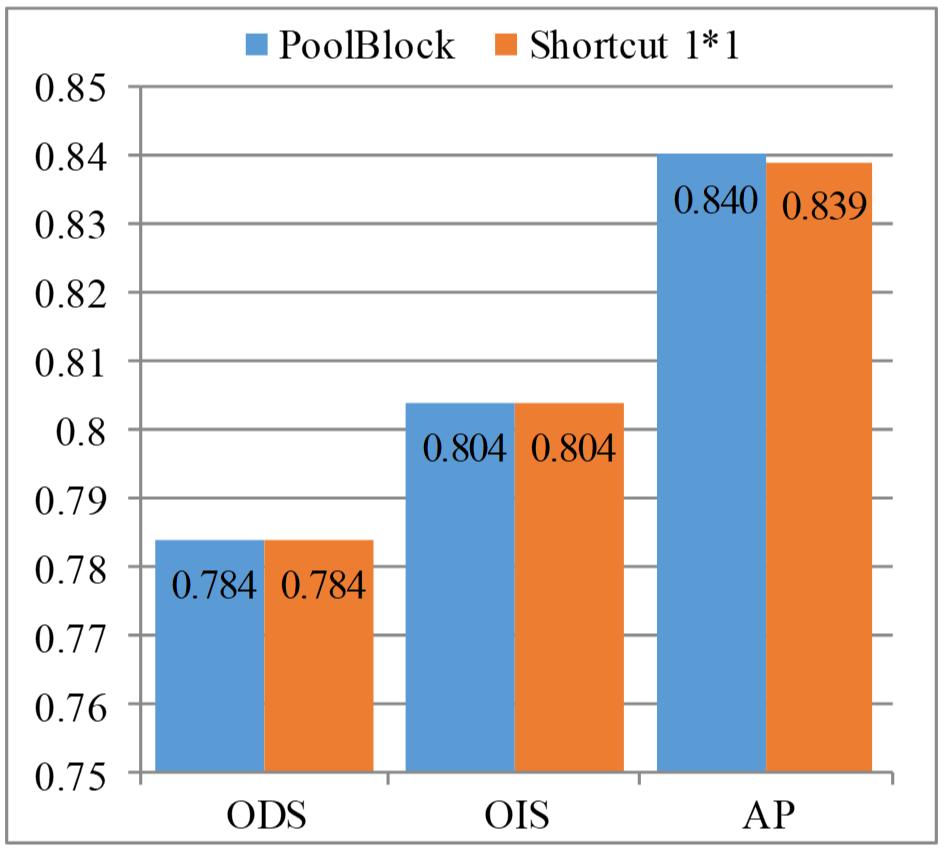}
	\caption{Comparison analysis of Shortcut 1×1 and PoolBlock.}\label{fig4}
\end{figure}

In ResNet \cite{he2016deep}, the network is divided into multiple stages, each containing a series of residual blocks. For the first residual block between adjacent stages, a 1×1 convolution layer is typically used to adjust the input channels to ensure dimension matching for shortcut connections. In lightweight network design, the goal is usually to reduce parameter count or improve computational efficiency, but using a 1×1 convolution layer to adjust input channels increases the parameter count. As shown in Fig. \ref{fig4}, using a 1×1 convolution layer (Shortcut 1×1) increased the parameters without surpassing the performance of our proposed PoolBlock. Therefore, in lightweight network design for edge detection, it is effective to omit the use of a 1×1 convolution layer to adjust the channels for the first residual block between adjacent stages.

\begin{table}[!ht]
	\centering
	\caption{Comparison analysis of different feature fusion methods. "$\times$" indicates no FBlock processing, "$\checkmark$" indicates FBlock processing.}
	\begin{tabular}{ccccc}
		\hline
		\textbf{X1} & \textbf{X2} & \textbf{ODS} & \textbf{OIS} & \textbf{AP} \\ \hline
		$\times$ & $\checkmark$ & .784 & .804 & .840  \\ 
		$\checkmark$ & $\checkmark$ & .784 & .805 & .840  \\ \hline
	\end{tabular}
	\label{table2}
\end{table}

Feature Fusion aims to integrate different feature information into an effective feature representation to enhance the model's performance and understanding. How to fuse the output features of different stages in the backbone network is a critical element. For the output features of adjacent stages, we tested two different feature fusion methods. As shown in Table \ref{table2}, X1 is the output feature of the previous stage, and X2 is the output feature of the next stage. The output feature X2 needs to be processed by the FBlock before feature fusion with X1. The experimental results in Table \ref{table2} show that processing only the output feature X2 of the subsequent stage with FBlock achieves performance comparable to processing both X1 and X2 with FBlock. Therefore, in lightweight network design, not processing the output feature X1 of the previous stage can reduce the model parameters without significant performance loss.

\subsection{Comparison with Other Models}
\label{subsec43}
The proposed method aims to achieve parameter efficiency and substantial detection performance. To evaluate its effectiveness, we compared it with two types of models: non-lightweight methods and lightweight methods. Non-lightweight methods include: HED \cite{xie2015holistically}, RCF \cite{liu2017richer}, CED \cite{wang2017deep}, DRNet \cite{cao2020learning}, LRNet \cite{lin2020lateral}, BDCN \cite{he2019bi}, CATS \cite{huan2021unmixing}, DexiNed \cite{soria2023dense}, EDTER \cite{pu2022edter}, DPED \cite{chen2022dped}, CHRNet \cite{elharrouss2023refined}. Lightweight methods include: PiDiNet \cite{su2021pixel}, TIN2 \cite{wibisono2020traditional}, FINED \cite{wibisono2021fined}, BDCN2 \cite{he2019bi}, BLEDNet \cite{luo2023blednet}, XYW-Net \cite{pang2024bio}.

In this paper, we present three different versions of experimental results: UHNet, UHNet-M, and UHNet-L. Their parameter counts increase sequentially, determined by the number of channels in the backbone network's different stages: UHNet has 32, 32, 32 channels in its three stages; UHNet-M has 32, 64, 128 channels; and UHNet-L has 64, 128, 256 channels.

\begin{figure}[!ht]
	\centering
	\includegraphics[width=1.0\textwidth]{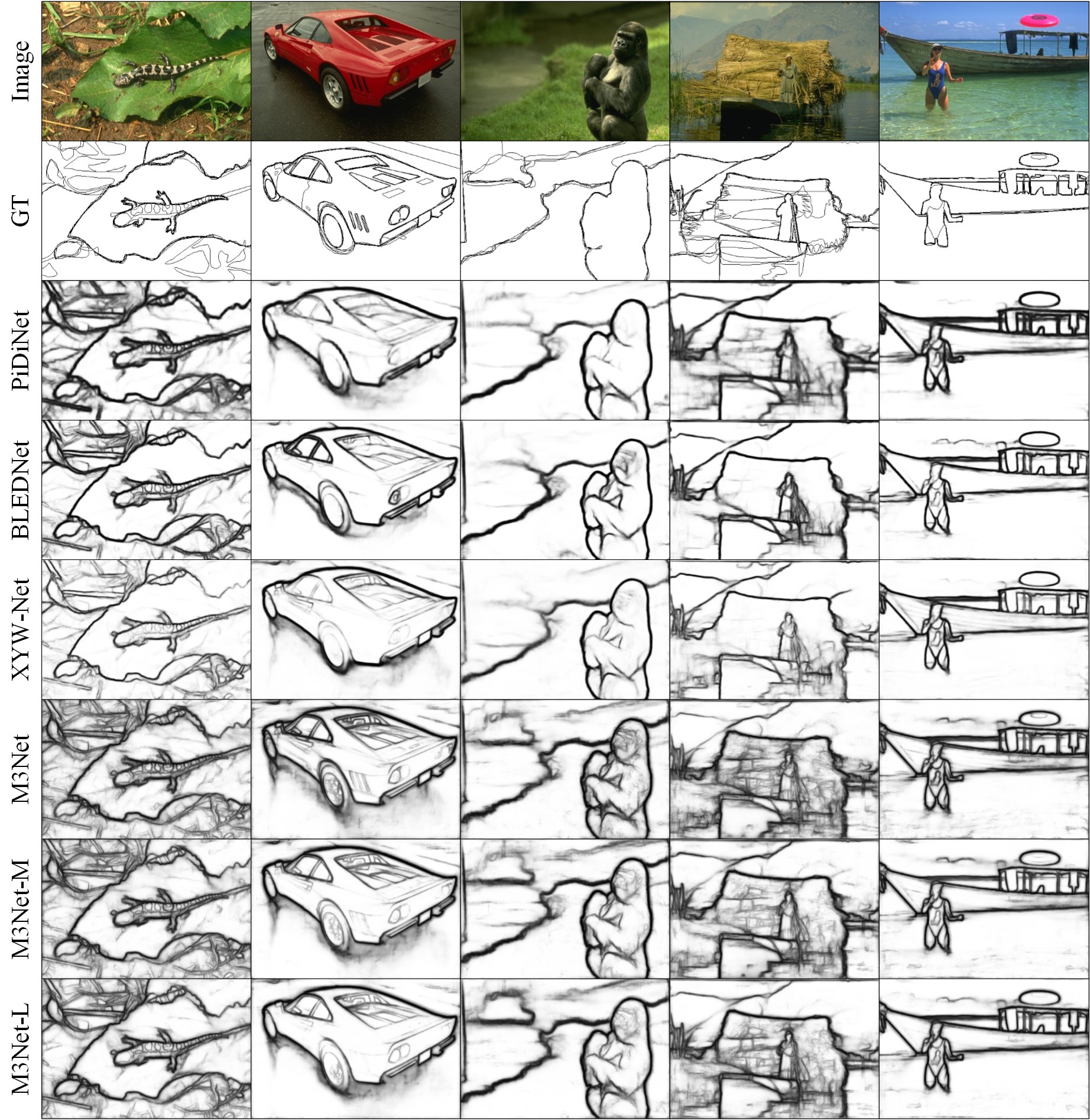}
	\caption{Qualitative comparison of the UHNet series models with other methods.}\label{fig5}
\end{figure}

\begin{table}[!ht]
	\centering
	\caption{Comparison with other methods on BSDS500 dataset.}
	\begin{tabular}{cccc}
		\hline
		\textbf{Method} & \textbf{ODS} & \textbf{OIS} & \textbf{AP} \\ \hline
		HED \cite{xie2015holistically}  & .788 & .808 & .840  \\ 
		RCF \cite{liu2017richer} & .806 & .823 & -  \\ 
		BDCN \cite{he2019bi} & .828 & .844 & .890  \\ 
		DRNet \cite{cao2020learning} & .817 & .832 & .836  \\ 
		LRNet \cite{lin2020lateral} & .820 & .838 & .874  \\ 
		EDTER \cite{pu2022edter} & .832 & .847 & .886  \\ 
		EPED \cite{chen2022dped} & .823 & .840 & .832  \\ 
		CHRNet \cite{elharrouss2023refined} & .787 & .788 & -  \\ 
		TIN2 \cite{wibisono2020traditional} & .772 & .795 & -  \\ 
		FINED \cite{wibisono2021fined} & .790 & .808 & -  \\ 
		PiDiNet-Baseline \cite{su2021pixel} & .798 & .816 & -  \\ 
		PiDiNet \cite{su2021pixel} & .807 & .823 & -  \\ 
		PiDiNet-Tiny \cite{su2021pixel} & .789 & .806 & -  \\ 
		PiDiNet-Tiny-L \cite{su2021pixel} & .787 & .804 & -  \\ 
		BDCN2 \cite{he2019bi} & .766 & .787 & -  \\ 
		BLEDNet \cite{luo2023blednet} & .805 & .823 & .851  \\ 
		XYW-Net \cite{pang2024bio} & .812 & .827 & .873  \\ 
		\textbf{UHNet} & .784 & .804 & .840  \\ 
		\textbf{UHNet-M} & .793 & .814 & .849  \\ 
		\textbf{UHNet-L} & .798 & .818 & .840  \\ \hline
	\end{tabular}
	\label{table3}
\end{table}

\begin{table}[!ht]
	\centering
	\caption{Comparison with other methods on NYUD dataset.}
	\label{table4}
	\begin{tabular}{ccccc}
		\hline
		\textbf{Method} & \textbf{Input} & \textbf{ODS} & \textbf{OIS} & \textbf{AP} \\ \hline
		\multirow{3}{*}{HED{ \cite{xie2015holistically}}} & RGB & .717 & .732 & .704 \\
		& HHA & .681 & .685 & .674 \\
		& RGB-HHA & .741 & .757 & .749 \\ \hline
		\multirow{3}{*}{BDCN{ \cite{he2019bi}}} & RGB & .748 & .763 & .770 \\
		& HHA & .707 & .719 & .731 \\
		& RGB-HHA & .765 & .781 & .813 \\ \hline
		\multirow{3}{*}{TIN2{ \cite{wibisono2020traditional}}} & RGB & .729 & .745 & - \\
		& HHA & .705 & .722 & - \\
		& RGB-HHA & .753 & .773 & - \\ \hline
		\multirow{3}{*}{PiDiNet-Tiny{ \cite{su2021pixel}}} & RGB & .721 & .736 & - \\
		& HHA & .700 & .714 & - \\
		& RGB-HHA & .745 & .763 & - \\ \hline
		\multirow{3}{*}{PiDiNet-Tiny-L{ \cite{su2021pixel}}} & RGB & .714 & .729 & - \\
		& HHA & .693 & .706 & - \\
		& RGB-HHA & .741 & .759 & - \\ \hline
		\multirow{3}{*}{CHRNet{ \cite{elharrouss2023refined}}} & RGB & .730 & .737 & - \\
		& HHA & .710 & .719 & - \\
		& RGB-HHA & .757 & .769 & - \\ \hline
		\multirow{3}{*}{BLEDNet{ \cite{luo2023blednet}}} & RGB & .730 & .747 & .716 \\
		& HHA & .710 & .728 & .698 \\
		& RGB-HHA & .757 & .775 & .772 \\ \hline
		\multirow{3}{*}{XYW-Net{ \cite{pang2024bio}}} & RGB & .730 & .747 & - \\
		& HHA & .701 & .715 & - \\
		& RGB-HHA & .756 & .775 & - \\ \hline
		\multirow{3}{*}{\textbf{UHNet}} & RGB & .694 & .713 & .683 \\
		& HHA & .681 & .699 & .671 \\
		& RGB-HHA & .727 & .747 & .740 \\ \hline
		\multirow{3}{*}{\textbf{UHNet-m}} & RGB & .704 & .721 & .693 \\
		& HHA & .690 & .706 & .674 \\
		& RGB-HHA & .734 & .753 & .745 \\ \hline
		\multirow{3}{*}{\textbf{UHNet-L}} & RGB & .707 & .724 & .698 \\
		& HHA & .692 & .709 & .683 \\
		& RGB-HHA & .735 & .755 & .748 \\ \hline
	\end{tabular}
\end{table}

\begin{table}[!ht]
	\centering
	\caption{Comparison with other methods on BIPED dataset.}
	\begin{tabular}{cccc}
		\hline
		\textbf{Method} & \textbf{ODS} & \textbf{OIS} & \textbf{AP } \\ \hline
		HED \cite{xie2015holistically} & .829 & .847 & .869  \\ 
		RCF \cite{liu2017richer} & .849 & .861 & .906  \\ 
		CED \cite{wang2017deep} & .795 & .815 & .830  \\ 
		BDCN \cite{he2019bi} & .890 & .899 & .934  \\ 
		CATS \cite{huan2021unmixing} & .887 & .892 & .817  \\ 
		DexiNed-f \cite{soria2023dense} & .895 & .900 & .927  \\ 
		DexiNed-a \cite{soria2023dense} & .893 & .897 & .940  \\ 
		XYW-Net \cite{pang2024bio} & .887 & .896 & .925  \\ 
		\textbf{UHNet} & .882 & .894 & .931  \\ 
		\textbf{UHNet-M} & .889 & .900 & .926  \\ 
		\textbf{UHNet-L} & .892 & .903 & .910  \\ \hline
	\end{tabular}
	\label{table5}
\end{table}

To verify the performance of the three different versions of UHNet, we conducted quantitative and qualitative experiments on the BSDS500 \cite{arbelaez2010contour}, NYUD \cite{silberman2012indoor}, and BIPED \cite{poma2020dense} datasets. The experimental results are given in Tables \ref{table3}, \ref{table4}, and \ref{table5}, and compared with other network models. The experimental data shows that UHNet has the smallest parameter count among all current deep learning-based algorithms (Table \ref{table6}), with only 42.3k, nearly 42\% less than PiDiNet-Tiny-L's \cite{su2021pixel} 73.0k, and achieves comparable performance with PiDiNet-Tiny-L \cite{su2021pixel} in the OIS metric on the BSDS500 \cite{arbelaez2010contour} dataset. Additionally, on the BIPED \cite{poma2020dense} dataset, UHNet-M, with 232.9k parameters, significantly outperforms XYW-Net's \cite{pang2024bio} 0.79M parameters. Notably, the ultra-lightweight version of UHNet is highly competitive in the OIS and AP metrics compared to other methods, both lightweight and non-lightweight. Although UHNet's performance on the NYUD \cite{silberman2012indoor} dataset is slightly inferior to other models, especially the state-of-the-art models, overall, our proposed UHNet detection model remains sufficiently competitive in terms of parameter count and computational complexity on the BSDS500 \cite{arbelaez2010contour} and BIPED \cite{poma2020dense} datasets, which is the focus of our research. Fig. \ref{fig5} presents the qualitative detection results of the UHNet series models compared to other methods.

\begin{table}[!ht]
	\centering
	\caption{Comparison of different methods in FLOPs and FPS performance metrics. Calculated based on 200×200 images.}
	\label{table6}
	\begin{tabular}{ccccc}
		\hline
		\textbf{Method} & HED \cite{xie2015holistically} & CED \cite{wang2017deep} & BDCN \cite{he2019bi} & DPED \cite{chen2022dped} \\ \hline
		\textbf{Params} & 14.7M & 21.8M & 16.3M & 67.9M \\
		\textbf{FLOPs(G)} & 24.3 & 60.8 & 37.0 & 63.4 \\
		\textbf{FPS} & 54 & 22 & 23 & 5 \\ \hline
		\textbf{Method} & FINED \cite{wibisono2021fined} & TIN2 \cite{wibisono2020traditional} & PiDiNet \cite{su2021pixel} & BLEDNet \cite{luo2023blednet} \\ \hline
		\textbf{Params} & 1.4M & 240k & 720.0k & 440.0k \\
		\textbf{FLOPs(G)} & 22.8 & 10.0 & 6.5 & 3.4 \\
		\textbf{FPS} & 26 & 47 & 42 & 50 \\ \hline
		\textbf{Method} & XYW-Net \cite{pang2024bio} & \textbf{UHNet} & \textbf{UHNet-M} & \textbf{UHNet-L} \\ \hline
		\textbf{Params} & 790.0k & 42.3k & 232.9k & 873.4k \\
		\textbf{FLOPs(G)} & 6.3 & 0.79 & 1.83 & 6.69 \\
		\textbf{FPS} & 27 & 166 & 151 & 136 \\ \hline
	\end{tabular}
\end{table}

We present in Table \ref{table6} the comparison of different methods on model size (Params), FLOPs, and FPS performance metrics. These experimental data were calculated based on 200×200 input images. The data shows that our proposed UHNet series detection models achieve high FPS and low FLOPs, particularly UHNet, which, while maintaining competitive ODS, OIS, and AP performance metrics, achieves 0.79 FLOPs and 166 FPS, significantly outperforming all other methods.

\section{Conclusion and Discussion}
\label{sec5}
This paper proposes an ultra-lightweight edge detection model with minimal parameters and fast speed. The smallest version of UHNet demonstrates strong competitiveness on the BSDS500 \cite{arbelaez2010contour} and BIPED \cite{poma2020dense} datasets with 42.3k parameters, 166 FPS, and 0.79G FLOPs. Our contributions to lightweight edge detection model design are threefold: first, based on the Bottleneck structure in the ResNet network \cite{he2016deep}, we propose an ultra-lightweight feature extraction module (PDDP block) with advantages such as fewer parameters, scalability, and transferability. Second, we optimized the residual connection between different stages in the backbone network, eliminating the need for 1×1 convolutions for channel transformation, overcoming this limitation. Third, for the output features of different stages, we explored more lightweight feature fusion methods that achieve comparable performance with other feature fusion methods while further reducing parameter count. We conducted extensive edge detection experiments on the BSDS500 \cite{arbelaez2010contour}, NYUD \cite{silberman2012indoor}, and BIPED \cite{poma2020dense} datasets. We believe that the UHNet series models have very strong competitiveness in terms of accuracy and efficiency.

Additionally, this study highlights several promising directions for further exploration. First, integrating traditional edge detectors (e.g., PiDiNet \cite{su2021pixel}) or drawing on biological visual physiological mechanisms (e.g., BLEDNet \cite{luo2023blednet}, XYW-Net \cite{pang2024bio}) with convolutional neural networks can achieve robust and accurate edge detection. Second, the loss functions used for edge detection tasks are almost all borrowed from other computer vision tasks, with certain limitations. Future work can explore loss functions more suitable for edge detection tasks to extract effective target edge information from abundant texture information. Third, the combination of lightweight network design and training with limited data samples from scratch is a key research focus in edge detection. Exploring more efficient lightweight network models based on lightweight network design and training with limited data samples, and drawing on efficient methods from other computer vision tasks (e.g., lightweight ViT \cite{han2022survey,khan2022transformers}, Mamba \cite{gu2023mamba,zhu2024vision,yu2024mambaout}), is also a worthwhile direction for research.

We hope that the proposal of UHNet provides more and newer insights into the design of ultra-lightweight edge detection network models. We also believe that efficient edge detection and extraction will play an even more significant role in engineering applications, particularly in medical image processing, as target edges serve as the foundation for other advanced visual tasks.

\section*{CRediT Authorship Contribution Statement}
Fuzhang Li: Conceptualization, Methodology, Validation, Funding acquisition, Writing – original draft, Writing – review and editing. Chuan Lin: Conceptualization, Writing – review and editing, Funding acquisition.

\section*{Declaration of Interest Statement}
The authors declare that they have no known competing financial interests or personal relationships that could have appeared to influence the work reported in this paper.

\section*{Data Availability}
Data will be made available on request.

\section*{Acknowledgement}
This work was supported by The Basic Ability Enhancement Program for Young and Middle-aged Teachers of Guangxi (2023KY0356), National Natural Science Foundation of China (Grant No.62266006), Guangxi Natural Science Foundation (Grant No.2020GXNSFDA297006) and National Natural Science Foundation of China (Grant No.61866002).

%
%
%


\end{document}